%% file: main.tex
\documentclass[11pt,a4paper]{article}
\usepackage[utf8]{inputenc}
\usepackage[hyperref]{acl2021}
\usepackage{times}
\usepackage{latexsym}

\usepackage{microtype}
\usepackage{graphicx}

\usepackage{amsmath}

\usepackage{booktabs}
\usepackage{enumitem}
\usepackage{float}
\usepackage{hyperref}
\usepackage{adjustbox}

\usepackage{makecell}
\usepackage{multirow}
\usepackage{colortbl}

\aclfinalcopy 
\def\aclpaperid{11} 

\usepackage{soul}
\usepackage{mysymbols}

\title{Comprehension Based Question Answering using Bloom's Taxonomy}

\makeatletter
\newcommand\footnoteref[1]{\protected@xdef\@thefnmark{\ref{#1}}\@footnotemark}
\makeatother

\author{
Pritish Sahu$^{1,2}$ \thanks{These two authors contributed equally.} \quad
Michael Cogswell$^{1\;*}$ \quad
Sara Rutherford-Quach$^{1}$ \quad
Ajay Divakaran$^{1}$\\
$^{1}$SRI International \\
$^{2}$Rutgers University \\
\\
}

\date{}

\begin{document}
\maketitle
\input{abstract}

\input{intro}

\input{related_work}

\input{approach}

\input{exp}

\input{conclusion}

\section*{Acknowledgments}
The authors thank Yunye Gong, Stephanie Nunn, and the anonymous reviewers for
the helpful discussions and comments.

\bibliographystyle{acl_natbib}
\bibliography{acl2021}

\appendix

\input{appendix}

\end{document}

%% file: abstract.tex
\begin{abstract}
Current pre-trained language models have lots of 
knowledge, but a more limited ability to use that knowledge.
Bloom's Taxonomy helps educators teach children how to use knowledge
by categorizing comprehension skills, so we use 
it to analyze and improve the comprehension skills of large 
pre-trained language models.
Our experiments focus on zero-shot question answering,
using the taxonomy to provide proximal context that helps the model
answer questions by being relevant to those questions.
We show targeting context in this manner improves performance
across 4 popular common sense question answer datasets.

\end{abstract}

%% file: intro.tex
\section{Introduction}
\label{sec:intro}

Recent large language models
such as GPT-3~\cite{gpt3} have made a giant leap forward in knowledge acquisition
and even generalize this knowledge to a
new tasks.
But when less narrow tasks are considered
they fail to understand as much as these benchmarks suggest.
They turn out to be  ``stochastic parrots''~\cite{stochastic_parrots} or ``smart/super parrots.''~\cite{define_comprehension} that just memorize without
all of the comprehension we want from a Natural Language Understanding system.
We focus on a particular kind of failure mode where the model knows (has memorized)
the information it needs, but is not able to apply that information correctly,
and we do so in a zero-shot fashion to control for what the model knows.

\begin{figure*}[t]
    \begin{center}
       \includegraphics[width=0.9\linewidth]{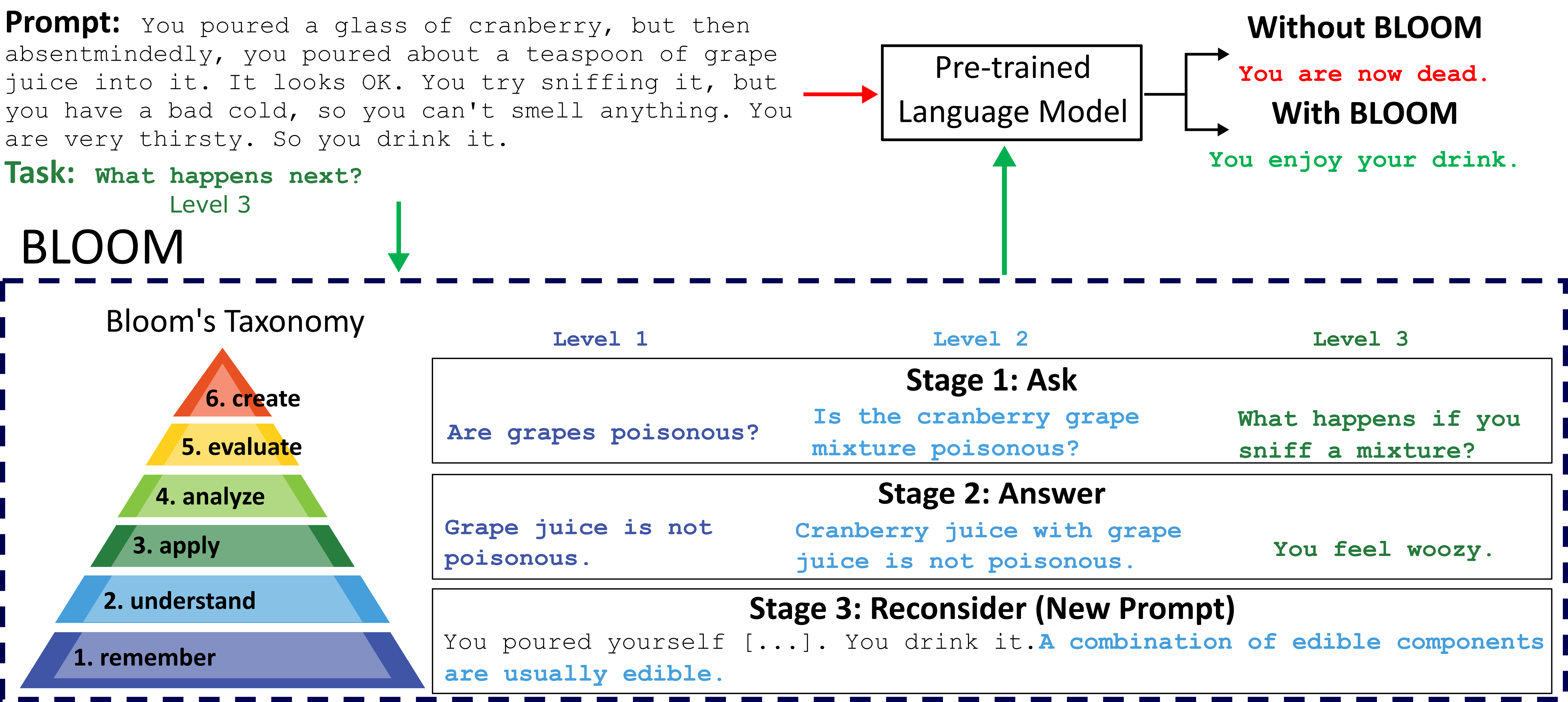}
    \end{center}
    \caption{Our approach incorporates context into question answering guided by Bloom's Taxonomy.}
    \label{fig:teaser}
\end{figure*}

For example, in \figref{fig:teaser} the model is asked
if a mixture of grape juice and cranberry juice is safe to drink~\cite{marcus_bloviator}.
GPT-3 declares that it is a deadly poison, even though it
appears to "know" that grape juice and cranberry juice are safe to drink
by themselves (\figref{fig:teaser}, Level 1, dark purple). It even knows that cranberry juice with grape juice is
not poisonous, but it still thinks the result is death (\figref{fig:teaser}, Level 2, light blue).
The model has memorized the necessary information from large amounts of text, but does
not use its knowledge appropriately. Following \cite{shwartz2020unsupervised},
we extract this knowledge as explicit language then feed it back as additional
context during inference, forcing the model to use what it already knows but
in our case targeting specifically useful knowledge.


To formalize this distinction we drew inspiration from elementary school
classrooms, 
where teachers~\cite{miller2002reading,harvey2007strategies} have a schema based approach in
which they teach children to demonstrate multiple levels of comprehension,
making complex inferences and direct recall from memory.
They use a hierarchy of comprehension skills called
Bloom's Taxonomy~\cite{bloomv2} (\cf \figref{fig:teaser}) with
memorization is at the bottom (requiring children to recall  facts)
followed by understanding (requiring children to grasp semantics)
application (requiring children to solve problems),
and more complex skills.
For us, these comprehension skills describe ways our language model
might fail to use its knowledge.

In this paper we address our failure mode by relying on commonly understood relationships between
the skills of Bloom's Taxonomy which we term \emph{proximal context}.
In order to \emph{understand} whether the cranberry grape
mixture is poisonous the model needs to \emph{remember} whether grape juice is poisonous.
In order to \emph{apply} its knowledge to figure out what will happen next it needs to
\emph{understand} whether the cranberry grape mixture is poisonous or not.
In general, the \emph{proximal context} for a particular task $T$ at level $L$ is given by
those tasks implicitly required by $T$, which are mostly at level $L-1$ of the taxonomy.
We guide our language to answer questions more accurately
by providing it not just any context, but proximal context
\footnote{Proximal context is not defined for level 1 questions,
so we only address questions at level 2 or above.}.
In performing zero-shot question answering our language model asks itself
additional clarification questions, choosing those most likely to result
in proximal context.

Our contributions in this paper are:
\begin{itemize}
    \item 
    We use Bloom's Taxonomy to choose proximal clarifying context that improves question answering performance using only what the model already knows.
    
    \item
    We show proximal context is better than other levels of context on four different commonsense question answering tasks.
    
    \item
    By observing how different levels of clarification impact our language model we 
    also explain how the model answers questions.
\end{itemize}

%% file: related_work.tex
\section{Related Works}

\para{Question Answering from External Supervision}
Several approaches has been proposed to improve question-answering by adding external knowledge source. Recent large pre-trained language models \cite{peters2018deep,radford2019language,devlin2018bert,liu2019roberta,joshi2020contextualized,clark2020electra} learn general purpose text encoders from a huge text corpus. \cite{petroni2019language} recently used a language model as knowledge
base to unmask a token given an entity and a relation in a predefined template. \citet{shwartz2020unsupervised,bosselut2019dynamic,bosselut2019comet} used pretrained language models to improve zero-shot question answering performance by extracting
context from the language model itself, using self-talk or a knowledge graph.
We add context via self-talk, with structure provided by Bloom's Taxonomy.

\para{Bloom's Taxonomy}
The original work~\cite{bloomv1} defined taxonomies for learning in the cognitive 
(intellectual), affective (interests, attitudes, values), and psychomotor domains, 
though the cognitive domain is what we usually refer to today.
Almost half a century later the cognitive domain taxonomy was revised~\cite{bloomv2}
to reflect more active thinking and improve usability by adding verbs to describe levels.
Teachers use this taxonomy, for example in computer science education~\cite{Whalley2006AnAS,Thompson2008BloomsTF,Oliver2004ThisCH},
and our inspiration is from this revised version of the cognitive taxonomy.
Machine learning has been applied to automatically classify questions~\cite{Mohammed2020QuestionCB,Zhang2021AutomatedCO,Nafa2016AutomaticCC}
into Bloom's Taxonomy levels, but the taxonomy has not been applied
to analyze or improve machine learning models themselves.
We use it to help our model think about what it knows.


%% file: approach.tex
\section{Approach}
Our approach builds on the zero-shot question answering approach of
\citet{shwartz2020unsupervised} to answer questions (\secref{sec:qa})
by adding clarifications with self-talk (\secref{sec:selftalk}).
We describe this approach then we use Bloom's Taxonomy
to select better clarifications (\secref{sec:bloom_filter}).

\begin{table*}[!htbp]
\adjustbox{max width=\textwidth}{%
\centering
\begin{tabular}{llll}
\rowcolor[rgb]{0.753,0.753,0.753} \multicolumn{1}{c}{\textbf{Dataset}} & \multicolumn{1}{c}{\textbf{Question Prefix}} & \multicolumn{1}{c}{\begin{tabular}[c]{@{}>{\cellcolor[rgb]{0.753,0.753,0.753}}c@{}}\textbf{Sample Clarification}\\\textbf{ Question}\end{tabular}} & \multicolumn{1}{c}{\begin{tabular}[c]{@{}>{\cellcolor[rgb]{0.753,0.753,0.753}}c@{}}\textbf{Sample Clarification}\\\textbf{ Answer}\end{tabular}}  \\ 
\midrule
\multirow{2}{*}{\textbf{COPA}}   & What is the definition of : 1~     & \begin{tabular}[c]{@{}l@{}}(a) What is the definition of an accident?\\(b)~What is the definition of a flat tire?\end{tabular}    & \begin{tabular}[c]{@{}l@{}}(a) The definition of an accident is the crash of \\collusion caused by the vehicle.\\(b)~The definition of a flat tire is that the tire does\\~not hold air.\end{tabular}   \\ & 
What is the main purpose of : 2    & \begin{tabular}[c]{@{}l@{}}(a) What is the main purpose of this\\~investigation? \\(b)~What is the main purpose of this\\~post?\end{tabular}       & \begin{tabular}[c]{@{}l@{}}(a) The purpose of this investigation is to provide \\ information about how and why he was shot. \\(b)~The purpose of this post is to share my thoughts\\~and feelings on his death.\end{tabular}                \\ 
\midrule
\multirow{2}{*}{\textbf{CommonsenseQA}}        & What is the main function of a~: 2~          & \begin{tabular}[c]{@{}l@{}}(a)~What~is~the~main~function~of~a\\~teacher~in~this~area?\\(b)~What~is~the~main~function~of~a\\~farmer?\end{tabular}   & \begin{tabular}[c]{@{}l@{}}(a)The~main~function~of~a~teacher~in~this~area~is~to~\\teach~them~about~life~and~love.\\(b)~The~main~function~of~a~farmer~is~to~provide~food\\~for~his~family~and~the~community.\end{tabular}    \\   & What might have caused : 3                   & \begin{tabular}[c]{@{}l@{}}(a) What might have caused this problem? \\(b)~What~might~have~caused~the~animal~to~flee?\end{tabular}                  & \begin{tabular}[c]{@{}l@{}}(a) the cause of this problem was that his wife's \\ husband didn't have enough money. \\(b)~The~cause~of~the~animal~to~flee~was~a~predator.\end{tabular}                                                         \\ 
\midrule
\multirow{2}{*}{\textbf{Social IQA}}                                   & What did [NAME] do? : 1~                     & \begin{tabular}[c]{@{}l@{}}(a) What did Kendall do?\\(b) What did Kendall do?\end{tabular}                                                         & \begin{tabular}[c]{@{}l@{}}(a) What Kendall did was make sure that\\he wasn't going anywhere else.\\(b)~What~Kendall~did~was~so~horrible,~that\\~it~was~hard~to~believe.\end{tabular}  \\    & How would you describe [NAME]? : 3           & \begin{tabular}[c]{@{}l@{}}(a) How would you describe Riley? \\(b) How would you describe Riley?\end{tabular}                                      & \begin{tabular}[c]{@{}l@{}}(a) Riley is a big brother,~he's an awesome dad. \\(b)~Riley~is~a~very~sensitive~person~and~has~a~lot\\~of~anxiety.\end{tabular}  \\ 
\midrule
\multirow{2}{*}{\textbf{Winogrande}}                                   & What are the properties of a : 1             & \begin{tabular}[c]{@{}l@{}}(a) What are the properties of a diet \\that~is not healthy?\\(b) What are the properties of a home?\end{tabular}       & \begin{tabular}[c]{@{}l@{}}(a) The property of a diet that is not healthy\\are that it has high cholesterol (a good idea).\\(b) The properties of a home are that which\\~makes it comfortable and pleasant for the occupants.\end{tabular}  \\   & What does it mean to : 2                     & \begin{tabular}[c]{@{}l@{}}(a) What does it mean to be an explorer? \\(b) What does it mean to be sophisticated?\end{tabular}                      & \begin{tabular}[c]{@{}l@{}}(a) Be an explorer means to explore and make \\ sense of things. \\(b) Be sophisticated means to be classy, elegant\\~and smart.\end{tabular}                                                                     \\
\midrule
\end{tabular}
}
\caption{This table shows some of the question prefixes we used for
    different datasets in our experiments. 
    We assign each prefix a level in Bloom's Taxonomy. We show generated clarifications questions and answers for both Distil-GPT2 (a) and GPT-Neo (b) for their corresponding question prefixes.}
\label{table:bloom_sample}
\end{table*}

\subsection{Question Answering with Language Models}
\label{sec:qa}

Given a prompt $p$, a question $q$, and answer options $a_o \; \forall o \in [1, K]$
we use a pre-trained language model $LM$ to pick the correct answer $a_{o^*}$.
This approach simply concatenates each (prompt, question, answer) tuple into
into a single string of text $T_o = [p, q, a_o]$ and feeds this
string to the language model to assign each choice a score $s_o = LM(T_o)$.
The language model's answer is just the answer with
the highest score: $\hat{o} = \argmax_o s_o$.

\subsection{Self-talk Clarifications}
\label{sec:selftalk}
Self-talk~\cite{shwartz2020unsupervised} has a language model ask itself clarifying
questions then answer those questions to generate clarifications.

\para{Stage 1: Ask clarification questions}
To produce clarifications we start with a set of clarification question prefixes $r_1, \ldots, r_J$
that are designed specifically for each question answering dataset.
``What happens if'' is a sample prefix for the clarifications, shown in \figref{fig:teaser}, and in \tabref{table:bloom_sample}, we present examples for all the datasets we use.
In this stage the language model completes each of these prefixes, using its
generator function $LM_G$ to ask one question $R_j = LM_G(r_j)$ per prefix.

\para{Stage 2: Answer the questions}
Next we use the model to answer each of these questions, possibly prompted with an answer prefix $b_j$
corresponding to question prefix $r_j$. The results are the clarifications
$c_j = LM_G([R_j, b_j])$.

\para{Stage 3: Reconsider with a new prompt}
\label{sec:bloom_filter}
To use the clarifications we pick one from the list then append it to the original prompt.
This approach simply considers all combinations of clarifications questions and answers
$T_{j,o} = [p, q, c_{j}, a_o] \; \forall o,j$, first chooses the clarification
which maximizes model score per answer option, then chooses the final answer
$o^* = \argmax_{o} \max_j LM(T_{j, o})$.
This can improve question answering performance on its own, but
in the next section we more carefully choose clarifications using our notion of
proximal context and Bloom's Taxonomy.

\subsection{Using Bloom's Taxonomy to Choose Clarifications with Proximal Context}
To test our idea of proximal context we consider the level $L$ of task
give by each dataset then allow only proximal clarifications of level $L-1$.
We label each question prefix with the level of Bloom's Taxonomy that it falls into,
and then force the model to choose from the set 
$\mathcal{C}_L$ of clarifications of level $L$.
This results in a final choice for each level
$o_L^* = \argmax_{o} \max_{j \in \mathcal{C}_L}  LM(T_{j, o})$.
We also provide a \emph{Choice Baseline} that allows the model to choose any level of clarification
to show the model would have difficulty choosing proximal clarifications itself.
Note that the annotation of questions along Bloom's taxonomy requires special skills typically found only among educators. While a layperson can be trained to annotate such questions, our experience was that it takes much more time than we could afford for a preliminary study such as this one. We therefore relied on our co-author, Sara Rutherford-Quach, who is a researcher at SRI's Education Division and has also worked as a teacher at the kindergarten-elementary level to provide us the annotations. Two other co-authors, Sahu and Cogswell, went through those annotations and made sure that each label had a three way consensus among Rutherford-Quach, Sahu and Cogswell. 
There might be some ambiguity about which level a particular prefix fits into, but this is also true of other applications of the taxonomy~\cite{Thompson2008BloomsTF}.
In future work, we plan to carry out a more rigorous annotation with more than one skilled annotator so we can measure inter-annotator agreement through measures such as Kappa scores.


%% file: exp.tex
\section{Experiments}

\subsection{Datasets}
We evaluate our study on four datasets that can each be thought of in terms of multiple
choice question answering, all measuring some kind of common sense:
COPA~\cite{roemmele2011choice} measures common sense causal reasoning, 
CommonSenseQA~\cite{talmor2019commonsenseqa} asks questions that require prior knowledge,
Social IQA~\cite{sap2019social} asks about social common sense, 
and WinoGrande~\cite{sakaguchi2020winogrande} adversarially measures semantic common sense.
Perhaps surprisingly, all of the datasets we used asked questions that fell into just one level of the taxonomy (\tabref{tab:results}{}).
These datasets do focus on very specific problems,
but the result is still disappointing because it would be more useful to see variations
in both task and clarification level.
It may be interesting to develop
datasets that can better express the range of abilities described by Bloom's Taxonomy.

\begin{table}
\caption{Question answering accuracy and std. dev. using different levels ofclarification over multiple clarification samples. Results on the dev sets of each dataset.(* = level of proximal context \textbackslash{}wrt the dataset)}
\label{tab:results}
\small
\adjustbox{width=3.1in}{%
\centering
\begin{tabular}{llll} 
\hline
\textbf{Task}& 
\textbf{Model}& 
\textbf{Level}& 
\textbf{Accuracy}
\\ \midrule 
\multirow{3}{*}{
\begin{tabular}[c]{@{}l@{}}\\Winogrande\\ (1267 total)\\ (2: Understand)\end{tabular}} & 
\begin{tabular}[c]{@{}l@{}}Distil-GPT2 \\(235$\pm$5 valid)\end{tabular} &
\begin{tabular}[c]{@{}l@{}}0A: Choice Baseline\\ 1A: Remember*\\ 2A: Understand\end{tabular} &
\begin{tabular}[c]{@{}l@{}}53.2 $\pm$ 1.8\\ \textbf{54.7} $\pm$ \textbf{3.6} \\ 52.5 $\pm$ 3.1\end{tabular}                                         \\ &
\begin{tabular}[c]{@{}l@{}}GPT-Neo \\(1230$\pm$7 valid)\end{tabular} &
\begin{tabular}[c]{@{}l@{}}0A: Choice Baseline\\ 1A: Remember*\\ 2A: Understand\end{tabular} &
\begin{tabular}[c]{@{}l@{}}54.62 $\pm$ 0.5\\ \textbf{54.77} $\pm$ \textbf{0.5} \\ 54.76 $\pm$ 0.3\end{tabular}                                      \\ \midrule
\multirow{3}{*}{
\begin{tabular}[c]{@{}l@{}}\\SocialIQA\\ (1954 total)\\ (3: Apply)\end{tabular}} &
\begin{tabular}[c]{@{}l@{}}Distil-GPT2 \\(58$\pm$5 valid)\end{tabular} &
\begin{tabular}[c]{@{}l@{}}0B: Choice Baseline\\ 1B: Remember\\ 2B: Understand*\\ 3B: Apply\end{tabular} &
\begin{tabular}[c]{@{}l@{}}44.5 $\pm$ 0.1\\ 43.7 $\pm$ 2.1 \\ \textbf{48.0} $\pm$ \textbf{1.1} \\ 44.4 $\pm$ 1.8\end{tabular}                       \\ &
\begin{tabular}[c]{@{}l@{}}GPT-Neo \\(1334$\pm$9 valid)\end{tabular} &
\begin{tabular}[c]{@{}l@{}}0B: Choice Baseline\\ 1B: Remember\\ 2B: Understand*\\ 3B: Apply\end{tabular} & \begin{tabular}[c]{@{}l@{}}\textbf{48.74} $\pm$ \textbf{0.4} \\ 47.31 $\pm$ 0.1 \\ \textbf{48.44} $\pm$ \textbf{0.5} \\ 48.1 $\pm$ 0.1\end{tabular}                                    \\ \midrule
\multirow{2}{*}{
\begin{tabular}[c]{@{}l@{}}\\COPA\\ (100 total)\\ (3: Apply)\end{tabular}} &
\begin{tabular}[c]{@{}l@{}}Distil-GPT2 \\(11$\pm$2 valid)\end{tabular} &
\begin{tabular}[c]{@{}l@{}}0C: Choice Baseline\\ 1C: Remember\\ 2C: Understand*\\ 3C: Apply\end{tabular} & \begin{tabular}[c]{@{}l@{}}\textbf{54.9} $\pm$ \textbf{0.9}\\ 46.0 $\pm$ 14.7 \\ \textbf{53.1} $\pm$ \textbf{12.5} \\ 40.8 $\pm$ 15.2\end{tabular}  \\ &
\begin{tabular}[c]{@{}l@{}}GPT-Neo \\(96$\pm$0 valid)\end{tabular} &
\begin{tabular}[c]{@{}l@{}}0C: Choice Baseline\\ 1C: Remember\\ 2C: Understand*\\ 3C: Apply\end{tabular} & \begin{tabular}[c]{@{}l@{}}\textbf{70.83} $\pm$ \textbf{0.0}\\ 65.62 $\pm$ 0.0 \\ \textbf{70.83} $\pm$ \textbf{1.4} \\ \textbf{70.83} $\pm$ \textbf{0.0} \end{tabular}                           \\ \midrule
\multirow{3}{*}{\begin{tabular}[c]{@{}l@{}}\\CommonsenseQA\\ (1221 total)\\ (3: Apply)\end{tabular}} &
\begin{tabular}[c]{@{}l@{}}Distil-GPT2 \\(68$\pm$1 valid)\end{tabular} &
\begin{tabular}[c]{@{}l@{}}0D: Choice Baseline\\ 1D: Remember\\ 2D: Understand*\\ 3D: Apply\end{tabular} & \begin{tabular}[c]{@{}l@{}}\textbf{29.9} $\pm$ \textbf{2.7}\\ 26.5 $\pm$ 3.3 \\ \textbf{28.1} $\pm$ \textbf{1.2} \\ 25.6 $\pm$ 3.4\end{tabular}     \\ &
\begin{tabular}[c]{@{}l@{}}GPT-Neo \\(1118$\pm$4 valid)\end{tabular}  &
\begin{tabular}[c]{@{}l@{}}0D: Choice Baseline\\ 1D: Remember\\ 2D: Understand*\\ 3D: Apply\end{tabular} & \begin{tabular}[c]{@{}l@{}}40.59 $\pm$ 3.6\\ 38.00 $\pm$ 6.0 \\ \textbf{43.19} $\pm$ \textbf{0.2} \\ 42.30 $\pm$ 0.8\end{tabular}
\\ \midrule
\end{tabular}
}
\end{table}

\subsection{Language Model}
We use distill-GPT2~\cite{distill_transformers} and the publicly released GPT-Neo2.7B\cite{gpt-neo} (based on  EleutherAI's replication of the GPT-3 architecture) as the language models throughout our experiments.
Our clarification question prefixes and hyperparameter settings for both models are from \cite{shwartz2020unsupervised}.
For each question prefix, we generate 5 clarification questions using nucleus sampling threshold 
probability $p=0.2$ and adding at most 6 words to the clarification question prefix.
We then generate 10 answers to each clarification question using $p=0.5$ and maximum answer length 10.
Some changes were necessary
to accurately measure the impact of clarification level.
Instead of always including \emph{no clarification} as a choice we do not
allow this option as it defeats our goal of measuring clarification level impact.
Furthermore, we do not use the clarification questions which were manually completed
without input from the model (as in COPA and Winogrande).

In order to compare performance across different levels of clarifications we
only consider examples where the model was able to generate at
least one clarification from each level.
To increase the number of viable examples we found it necessary to remove some restrictions
relative to the implementation of \cite{shwartz2020unsupervised}.
In particular, we kept all clarifications that 
had no overlapping words with the context and did not allow the model to chose the
``no clarification'' option. Even with these constraints it was still often the case that
distil-GPT2 could not generate a short clarification sentence that was plausible
enough to use whereas GPT-Neo was able to generate clarifications for almost the entire dataset.
This indicates larger scale models may be more able to take advantage of clarifying questions.
The number of examples with valid clarifications for all levels is indicated for each model in column 2 of \tabref{tab:results}{}.
These changes help us more accurately measure the impact of Bloom's Taxonomy, but
mean our approach is not directly comparable to \citet{shwartz2020unsupervised}.

\subsection{Results}

Table \ref{tab:results} reports the performance of our Bloom’s Taxonomy
infused zero-shot question answering method. Each row shows question answering 
accuracy for a particular dataset and level of clarification. 
If our hypothesis is correct then the level of available clarifications should matter 
and clarifications that provide proximal context 
--one level below the dataset level-- should be most helpful.

\para{Clarification Level Makes a Difference}
All levels of clarification questions and answers provide some amount of extra
information that changes how a language model processes the entire string
it is presented with. This is often helpful information, but it may be that
all levels of Bloom's Taxonomy provide equally useful information.
We find that is not the case. Different levels of clarification help more or less,
as evidenced by the large gap between minimum and maximum accuracy
for each dataaset.
Furthermore, when the model can choose any clarification
(rows 0A/B/C/D) it either does a worse job than proximal context or its performance
similar to proximal context, so enforcing a particular kind of context should
be helpful.

\para{Proximal Context Helps Most}
Proximal context, as we've defined it with respect to Bloom's Taxonomy is
context from the clarification level directly below the dataset question level.
The proximal clarification level for each dataset is marked by a * in \tabref{tab:results}{}.
In all cases proximal clarifications are better than
using clarifications of a lower level. 
For the datasets that ask
level 3 questions the proximal (level 2) clarifications also outperform
level 1 clarifications (2B/C/D greater than 1B/C/D).
Proximal clarifications are also about as good as or better than
using clarifications of a higher level. You can see this for Winogrande
by noting row 1A is greater than 2A and for the other datasets by noting
rows 2B/C/D usually have greater performance than 3B/C/D.
Overall, proximal context is most consistent in efficacy.

\subsection{Qualitative Results}

In \tabref{table:bloom_sample}{} we show samples of question answer pairs generated
for each model and in \tabref{tab:full_examples}{} of the appendix we show
complete examples (with context and choices) for each model and dataset.
GPT-Neo is much larger than distil-GPT2 and is expected to generalize to
slightly new tasks like the clarification generation task better than the
smaller model. This expectation is clearly met by the observed quality of
clarifications. Distil-GPT2 clarification questions and answers often do
not have meaningful semantics, are not correct, or are not relevant.
GPT-Neo is much more likely to generate questions and answers which
are meaningful, correct, and relevant. 
This suggests the greater number of
valid clarifications generated by GPT-Neo may be due to an increase
in clarification quality.
Furthermore, it fails in an intuitive
fashion: when it fails to generate
meaningful answers it often has also failed to generate
a meaningful clarification question in the first place.

Also note that the performance differences observed for distil-GPT2 occur
despite its relatively poor interpretability.
This indicates that context which is somewhat relevant to the topic even if it
does not precisely make sense can still be useful.



%% file: conclusion.tex
\section{Conclusion}
\label{sec:conclusion}

Large pre-trained language models sometimes have the right information, but they
just do not know how to use it.
We used Bloom's taxonomy to pick questions with the right amount of proximal
context.
This helped the language models use their knowledge to more effectively answer questions.
In the future we would like to extend our work on tasks that present a wide range of questions that fall under different levels of the taxonomy. Similarly, we also would like to study and improve upon the current limited set of prefix questions used.

%% file: appendix.tex
\section{Prefixes and Examples}
In the appendix we provides more details about the question prefixes we used in \tabref{tab:prefix_level}{} and
provide more examples of outputs from our models in \tabref{tab:full_examples}{}.

\begin{table*}[!ht]
\caption{All the prefix questions with its corresponding taxonomy level used in our zero shot question answering evaluation.}
\centering
\adjustbox{max width=\textwidth}{%
\label{tab:prefix_level}
\begin{tabular}{llc} 
\hline
\textbf{Question Prefix}   & \textbf{Answer Prefix}  & \multicolumn{1}{l}{\textbf{Bloom's Taxonomy Level}}                               \\ 
\multicolumn{3}{c}{{\cellcolor[rgb]{0.753,0.753,0.753}}\textbf{CommonsenseQA \& COPA}}                                                    \\ 
\begin{tabular}[c]{@{}l@{}}What is the definition of\\ What is the main purpose of\\ What is the main function of a\\ What are the properties of a\\ What is a\\ What happened as a result of\\ What might have caused\end{tabular}  
& \begin{tabular}[c]{@{}l@{}}The definition of \_ is\\ The purpose of \_ is to\\ The main function of a \_ is\\ The properties of a \_ are that\\ \_ is\\ As a result of \_,\\ The cause of \_ was\end{tabular} 
& \begin{tabular}[c]{@{}c@{}}1\\2\\2\\1\\1\\3\\3\end{tabular}  
\\ 
\multicolumn{3}{c}{{\cellcolor[rgb]{0.753,0.753,0.753}}\textbf{SocialIQA }}
\\
\begin{tabular}[c]{@{}l@{}}What will [NAME] want to do next?\\What will [NAME] want to do after?\\How would [NAME] feel afterwards?\\How would [NAME] feel as a result?\\How would [NAME] feel after?\\How would you describe [NAME]?\\What kind of person is [NAME]?\\How would you describe [NAME] as a person?\\Why did [NAME] do that?\\Why did [NAME] do this?\\Why did [NAME] want to do this?\\What does [NAME] need to do beforehand?\\What does [NAME] need to do before?\\What does [NAME] need to do before this?\\What did [NAME] need to do before this?\\What will happen to [NAME]?\\What will happen to [NAME] next?\\What will [NAME] do next?\\What did [NAME] do?\end{tabular} & \begin{tabular}[c]{@{}l@{}}NAME] wanted\\{[}NAME] wanted\\{[}NAME] felt\\{[}NAME] felt\\{[}NAME] felt\\{[}NAME] is a\\{[}NAME] is a\\{[}NAME] is a\\{[}NAME] did this because they wanted\\{[}NAME] did this because they wanted\\{[}NAME] did this because they wanted\\Before doing that, [NAME] first had to\\Before doing that, [NAME] first had to\\Before doing that, [NAME] first had to\\Before doing that, [NAME] first had to\\{[}NAME]\\{[}NAME]\\{[}NAME]\\What [NAME] did was\end{tabular} &  \begin{tabular}[c]{@{}c@{}}3 \\ 3 \\ 3 \\ 3 \\ 3 \\ 2 \\ 2 \\ 2 \\ 3 \\ 3 \\ 3 \\ 2 \\ 2 \\ 2 \\ 2 \\ 3 \\ 3 \\ 3 \\ 1\end{tabular}  \\ \\
\multicolumn{3}{c}{{\cellcolor[rgb]{0.753,0.753,0.753}}\textbf{Winogrande }}
\\
\begin{tabular}[c]{@{}l@{}}What is the definition of \\ What is the main purpose of \\ What is the main function of a \\ What are the properties of a \\ What is \\ What does it mean to\end{tabular}  
& \begin{tabular}[c]{@{}l@{}}The definition of \_ is \\ The purpose of \_ is to \\ The main function of a \_ is \\ The properties of a \_ are that \\ \_ is \\ \_ means\end{tabular} 
& \begin{tabular}[c]{@{}c@{}}1\\2\\2\\1\\1\\2\end{tabular}                                                                             \\
\hline
\end{tabular}
}
\end{table*}

\begin{table*}[!ht]
\caption{Example contexts, questions, choices, clarification questions and clarification answers for each dataset. We present results for both Distil-GPT2 and GPT-Neo.}
\centering
\adjustbox{max width=\textwidth}{%
\refstepcounter{table}
\label{tab:full_examples}
\begin{tabular}{lll}
\rowcolor[rgb]{0.753,0.753,0.753} \multicolumn{1}{c|}{\textbf{COPA}} & \multicolumn{2}{c}{\textbf{Generated Clarifications}}  \\ \midrule
\textbf{(Context, Question, Choice)} & \textbf{Distil-GPT2} & \textbf{GPT-Neo} \\
\begin{tabular}[c]{@{}l@{}}C:The man turned on the faucet.\\Q: effect\\Ch: [The toilet filled with water.,~ \textbf{}\\\textbf{~~~~~~ Water flowed from the spout.}]\end{tabular}  & \begin{tabular}[c]{@{}l@{}}Q: What might have caused it to be so hot?\\A: The cause of it to be so hot was that he would \\have a big fire in his face and then immediately \\had an explosion, which killed all but one person.\end{tabular}                & \begin{tabular}[c]{@{}l@{}}Q: What happened as a result of that?\\A: As a result of that, he was forced to drink water.\end{tabular}                                             \\
\begin{tabular}[c]{@{}l@{}}C: The woman retired.\\Q: effect\\Ch: [\textbf{She received her pension.},~ \textbf{}\\\textbf{~~~~~~ }She paid off her mortgage.]\end{tabular}                                                                                                          & \begin{tabular}[c]{@{}l@{}}Q: What happened as a result of this?\\A: As a result of this, the government has announced\\ that it will pay Rs 7 crore to her family for compensation \\and legal expenses related thereto as part-time payments.\end{tabular} & \begin{tabular}[c]{@{}l@{}}Q: What happened as a result of her retirement?\\A: As a result of her retirement, she had been receiving \\a pension for life.\end{tabular}          \\
\begin{tabular}[c]{@{}l@{}}C: I wanted to conserve energy.\\Q: effect\\Ch: [I swept the floor in the unoccupied room.,~ \\\textbf{~~~~~~ I shut off the light in the unoccupied room.}]\end{tabular}                                                                                & \begin{tabular}[c]{@{}l@{}}Q: What is the main function of a plant?\\A: The main function of a plant is the growth\\and maintenance of nutrients, so that we can \\increase our food intake by 10\% or more per year, he said.\end{tabular} & \begin{tabular}[c]{@{}l@{}}Q: What might have caused this?\\A: The cause of this was the lack of electricity in my home.\end{tabular}  \\
&   &    \\
\rowcolor[rgb]{0.753,0.753,0.753} \multicolumn{1}{c|}{\textbf{Winogrande}}  & \multicolumn{2}{c}{\textbf{Generated Clarifications}}  \\ \midrule
\textbf{(Sentence, Context, Choice)}  & \textbf{Distil-GPT2}    & \textbf{GPT-Neo}  \\
\begin{tabular}[c]{@{}l@{}}S: The home that my parents had when I was\\ in school was a lot nicer than my house now \\because the \_ was sophisticated.\\C: home\\Ch: [\textbf{home}, house]\\\end{tabular}                                                                         & \begin{tabular}[c]{@{}l@{}}Q: What is the main function of a building?\\A: The main function of a building is to keep \\people happy and healthy.\end{tabular}                                                                                               & \begin{tabular}[c]{@{}l@{}}Q: What are the properties of a home?\\A: he properties of a home are that which \\makes it comfortable and pleasant for the occupants.\end{tabular}  \\
\begin{tabular}[c]{@{}l@{}}S: He had enough time between classes to go\\ to a cafe or to the library. He went to the \_ \\because his paper could wait.\\C: classes\\Ch: [\textbf{cafe}, library]\end{tabular}                                                                      & \begin{tabular}[c]{@{}l@{}}Q: What is the definition of a book?\\A: The definition of a book is an article\\ written in order for it be read.\end{tabular}                                                                                                   & \begin{tabular}[c]{@{}l@{}}Q: What is the definition of cafe?\\A: cafe is defined as a place where people\\ gather for refreshment.\end{tabular}                                 \\
\begin{tabular}[c]{@{}l@{}}S: They had to eat a lot to gain the strength\\they had lost and be able to work, the \_ was too much.\\C: strength\\Ch: [\textbf{work}, strength]\end{tabular}                                                                                          & \begin{tabular}[c]{@{}l@{}}Q: What is the main function of a good diet?\\A: The main function of a good diet is not for \\food or health but also on exercise.\end{tabular}                                                                                  & \begin{tabular}[c]{@{}l@{}}Q: What is the definition of strength?\\A: strength is defined as : having power\\ or ability to do something.\end{tabular}   \\ &  &  \\
\rowcolor[rgb]{0.753,0.753,0.753} \multicolumn{1}{c|}{\textbf{SocialIQA}}    & \multicolumn{2}{c}{\textbf{Generated Clarifications}}  \\ \midrule
\textbf{(Context, Question, Choice)}   & \textbf{Distil-GPT2}    & \textbf{GPT-Neo}   \\
\begin{tabular}[c]{@{}l@{}}C: Kendall got a new sports car and could not wait \\to show friends.\\Q: What will Kendall want to do next?\\Ch: [drive that sports car,\\\textbf{show off his new sports car},\\clean and wax her legs]\end{tabular}                                   & \begin{tabular}[c]{@{}l@{}}Q: Why did Kendall want to do this?\\A: Kendall did this because they wanted her, too.\end{tabular}                                                                                                                               & \begin{tabular}[c]{@{}l@{}}Q: What will Kendall want to do next?\\A: Kendall wanted to impress them with \\his new ride."\end{tabular}                                           \\
\begin{tabular}[c]{@{}l@{}}C: Kendall was really into house plants and often\\ pour water in to jars so the fluoride would evaporate out.\\Q: What will Kendall want to do next?\\Ch: [\textbf{watered a few of the plants},\\gather jars,\\watered a few of her pets]\end{tabular} & \begin{tabular}[c]{@{}l@{}}Q: Why did Kendall do that?\\A: Kendall did this because they wanted him at \\home, he said.\end{tabular}                                                                                                                         & \begin{tabular}[c]{@{}l@{}}Q: Why did Kendall do this?\\A: Kendall did this because they wanted to\\ get a lot of water for their plants.\end{tabular}                           \\
\begin{tabular}[c]{@{}l@{}}C: Remy gave birth to a baby girl on a Saturday afternoon.\\Q: What does Remy need to do before this?\\Ch: [leave her husband,\\did this to start a family,\\\textbf{go to the hospital}]\end{tabular}                                                   & \begin{tabular}[c]{@{}l@{}}Q: What did Remy need to do before this?\\A: Before doing that, Remy first had to tell her \\he was too old.\end{tabular}                                                                                        & \begin{tabular}[c]{@{}l@{}}Q: What does Remy need to do before?\\A: Before doing that, Remy first had to go to \\the hospital to deliver her baby.\end{tabular}                  \\
\rowcolor[rgb]{0.753,0.753,0.753} \multicolumn{1}{c|}{\textbf{CommonsenseQA}}  & \multicolumn{2}{c}{\textbf{Generated Clarifications}}  \\ \midrule
\textbf{(Context, Question, Choice)}   & \textbf{Distil-GPT2}    & \textbf{GPT-Neo}   \\
\begin{tabular}[c]{@{}l@{}}C: In what Spanish speaking North American country \\can you get a great cup of coffee?\\Q: cup of coffee\\Ch: [mildred's coffee shop, \textbf{mexico}, diner, \\kitchen, canteen]\end{tabular}                                                          & \begin{tabular}[c]{@{}l@{}}Q: What is the definition of coffee?\\A: The definition of coffee is the word \\that comes from Latin for espresso.\end{tabular}                                                                                                  & \begin{tabular}[c]{@{}l@{}}Q: What is the definition of mexico?\\A: mexico is defined as the country that \\produces the best coffee.\end{tabular}                               \\
\begin{tabular}[c]{@{}l@{}}C: Unlike a spider and his many sight seers, people \\only have what?\\Q: people\\Ch: [tongues, names, brains, feelings, \textbf{two eyes}]\end{tabular}                                                                                                 & \begin{tabular}[c]{@{}l@{}}Q: What is the main purpose of this article?\\A: The purpose of this article is to explain \\the basic characteristics that are characteristic of spiders.\end{tabular}                                                           & \begin{tabular}[c]{@{}l@{}}Q: What is the definition of two eyes?\\A: two eyes is defined as the organs \\by which we see.\end{tabular}                                          \\
\begin{tabular}[c]{@{}l@{}}C: The artist was sitting quietly pondering, then \\suddenly he began to paint when what struck him?\\Q: sitting quietly\\Ch: [sadness, anxiety, \textbf{inspiration}, discomfort, insights]\end{tabular}                                                         & \begin{tabular}[c]{@{}l@{}}Q: What is a man who can do that?\\A: Man who can do that is a really great painter.\end{tabular}                                                                                                                                 & \begin{tabular}[c]{@{}l@{}}Q: What is the definition of inspiration?\\A: inspiration is defined as a sudden flash \\of inspiration that comes from within.\end{tabular}          \\  &    &         
\end{tabular}
}
\end{table*}
\label{sec:appendix}